\documentclass[11pt]{article} 
\usepackage{algorithm}
\usepackage[noend]{algorithmic}

\usepackage{rldmsubmit,palatino}
\usepackage{graphicx}
\usepackage{microtype}
\usepackage{graphicx, caption}
\usepackage{subfigure}
\usepackage{booktabs, bm, enumitem} 
\usepackage{color,soul}

\usepackage[utf8]{inputenc} 
\usepackage[T1]{fontenc}    
\usepackage{url}            
\usepackage{booktabs}       
\usepackage{amsfonts}       
\usepackage{nicefrac}       
\usepackage{microtype}      
\usepackage{xcolor}         
\usepackage{natbib}

\usepackage{amsmath}
\usepackage{amssymb}
\usepackage{mathtools}
\usepackage{amsthm}

\usepackage{graphicx}
\usepackage{xcolor, amssymb, enumitem}
\usepackage[colorlinks=true,linkcolor=black,citecolor={rgb, 255:red, 160; green, 0; blue, 0 }]{hyperref}
\usepackage{titlesec}
\titlespacing*{\section}{0pt}{0.5em}{0.5em}
\titlespacing*{\subsection}{0pt}{0.5em}{0.5em}

\theoremstyle{plain}
\newtheorem{theorem}{Theorem}[section]
\newtheorem{proposition}[theorem]{Proposition}

\theoremstyle{definition}
\newtheorem{definition}[theorem]{Definition}

\theoremstyle{remark}

\title{Towards Causal Model-Based Policy Optimization}

\author{
    Alberto Caron\thanks{Corresponding author.} \\
    The Alan Turing Institute\\
    London, UK\\
    \texttt{acaron@turing.ac.uk} \\
\And
    Vasilios Mavroudis \\
    The Alan Turing Institute\\
    London, UK\\
    \texttt{vmavroudis@turing.ac.uk} \\
\And
    Chris Hicks \\
    The Alan Turing Institute\\
    London, UK\\
    \texttt{c.hicks@turing.ac.uk} \\
}

\begin{document}

\maketitle

\begin{abstract}
Real-world decision-making problems are often marked by complex, uncertain dynamics that can shift or break under changing conditions. Traditional Model-Based Reinforcement Learning (MBRL) approaches learn predictive models of environment dynamics from queried trajectories and then use these models to simulate rollouts for policy optimization. However, such methods do not account for the underlying causal mechanisms that govern the environment, and thus inadvertently capture spurious correlations, making them sensitive to distributional shifts and limiting their ability to generalize. The same naturally holds for model-free approaches. In this work, we introduce Causal Model-Based Policy Optimization (C-MBPO), a novel framework that integrates causal learning into the MBRL pipeline to achieve more robust, explainable, and generalizable policy learning algorithms.

Our approach centers on first inferring a Causal Markov Decision Process (C-MDP) by learning a local Structural Causal Model (SCM) of both the state and reward transition dynamics from trajectories gathered online. C-MDPs differ from classic MDPs in that we can decompose causal dependencies in the environment dynamics via specifying an associated Causal Bayesian Network. C-MDPs allow for targeted interventions and counterfactual reasoning, enabling the agent to distinguish between mere statistical correlations and causal relationships. The learned SCM is then used to simulate counterfactual on-policy transitions and rewards under hypothetical actions (or ``interventions"), thereby guiding policy optimization more effectively. The resulting policy learned by C-MBPO can be shown to be robust to a class of distributional shifts that affect spurious, non-causal relationships in the dynamics. We demonstrate this through some simple experiments involving near and far OOD dynamics drifts.

\end{abstract}


\startmain 

\section{Introduction}

Model-Based Reinforcement Learning (MBRL) offers a promising pathway towards sample-efficient policy learning by simulating agent-environment interactions from a learned model of the dynamics \citep{moerland2023model}. However, existing MBRL methods often rely on conventional predictive paradigms, which may fail to capture the underlying causal mechanisms. Consequently, learned policies lack explainability, can be brittle to distribution shifts, and may not effectively transfer to new tasks \citep{richens2024robust}. To address these challenges, we propose a novel Causal Model-Based Policy Optimization (C-MBPO) framework, which aims at solving the task best described by a Causal Markov Decision Process (MDP), which augments the classical MDP with a causal Bayesian network representing both reward and state-transition causal dynamics. By explicitly modeling and/or learning causal relationships, the agent can leverage counterfactual reasoning for more reliable policy learning that results in more robust, generalizable policies.

\vspace{-0.05cm}

\textbf{Related Work.} Foundational work by \cite{pearl2009causality} emphasizes the crucial role of representing, identifying, and exploiting causal relationships to enable robust decision-making in complex environments.  Several studies have adopted a causal perspective on the topic of explainability in RL \citep{madumal2020explainable}, where causal reasoning can enhance policy transparency. The field of ``causal RL'' has then gained momentum, offering diverse methods to integrate causal structures into RL for improved generalization and robustness \cite{zeng2024survey}. Researchers have explored the possibility of learning causal representations via targeted exploration \cite{sontakke2021causal}, exploiting invariant causal mechanisms in policy learning \citep{zhang2020invariant}, and leveraging prior causal knowledge for more efficient exploration \cite{lu2022efficient}. Other recent efforts have investigated frameworks that unify causal discovery and RL \cite{mendez2022causal}, detect causal influences for efficient training \cite{seitzer2021causal}, and design explainable RL agents through causal world models \cite{yu2023explainable}. Building on these foundations, our work contributes specifically to the intersection of causal learning and Model-Based Reinforcement Learning (MBRL).

\vspace{-0.05cm}

\textbf{Contributions.} We make two primary contributions. First, we formalize a novel framework of \emph{Causal Markov Decision Processes (C-MDPs)}, that augments standard MDPs with an explicit representation of the causal structure underlying state transitions and rewards. Second, we propose a new method, \emph{Causal Model-Based Policy Optimization (C-MBPO)}, which harnesses the C-MDP underlying a MDP task to guide a Dyna-Q style policy optimization \citep{janner2019trust}. Unlike existing approaches to MBRL \citep{moerland2023model}, our method jointly infers causal mechanisms and leverages structural interventions, yielding policies that remain robust under various distribution shifts \citep{richens2024robust}.

\section{Problem Setup}

Consider the problem defined by the classical MDP. An MDP is a tuple $\mathcal{M} = \langle {\mathcal{S}}, \mathcal{A}, p_{\tau}, p_0,r, \gamma  \rangle$, defined over an horizon of $t \in \{1, ..., T\}$ time steps where: i) $\mathcal{S}$ is the state space ; ii) $\mathcal{A}$ is the action space; iii) $p_{\tau} (s' | s,a) \in \mathcal{P}_{\tau}$ is the state transition probability; iv) $p_0 \in \mathcal{P}_0$ is the initial state probability function; v) $r : \mathcal{S} \times \mathcal{A} \rightarrow \mathcal{C} \subset \mathbb{R}$ is a bounded reward function; vi) $\gamma \in (0, 1)$ is a discount factor. The classic MDP formulation by itself does not incorporate any structural causal information about the environment's dynamics. In order to include such information, we need to resort to Causal Graphical Models and Causal Bayesian Networks \citep{pearl2009causality}. A Causal Bayesian Network (CBN) is a pair $\mathcal{B}= (\mathcal{G}, \Theta)$ where: i) $\mathcal{G}$ is a Causal Graphical Model (CGM), where vertices represent random variables $X_j \in \mathcal{V}$ and edges causal relationships $E_j \in \mathcal{E}$; ii) $\Theta$ is the space of parameters that characterizes the joint distribution over the random variables, $p_{\theta} (X_1, ..., X_d) = \prod^d_{i=1} p \big( X_i \mid pa(X_i) \big)$. In a CGM, $pa(X_i)$ denotes all the parents of $X_i$ (all the variables $X_j$ that have an inbound causal edge to $X_i$). An intervention on a variable is denoted as $do(X_i = x)$. Merging together the formalisms of MDPs and CBNs, we obtain a new object, a Causal Markov Decision Process (C-MDP), that can exhaustively describe causal relationships in the MDP dynamics. This is defined as:
\begin{definition}[Causal Markov Decision Process]
    A \textbf{Causal Markov Decision Process} (C-MDP) is a pair $\mathcal{M}_c = (\mathcal{M}, \mathcal{B}_{r, s'})$, where $\mathcal{M}$ is a standard MDP, while $\mathcal{B}_{r, s'} (s,a) = (\mathcal{G}_{r, s'} (s,a), \Phi)$ is the local Causal Bayesian Network underlying the probabilistic dynamics $p_{\theta}(r, s' \mid s,a)$. The local Causal Graphical Model $\mathcal{G}_{r, s'} (s,a)$ in $\mathcal{B}_{r, s'} (s,a)$ encodes all the inbound causal edges to $(r,s')$, from $pa(r,s')$.
\end{definition}
By construction of the original MDP problem, there is no same-time dependency between $r_t$ and $s_{t+1}$, or in other words, $r_t$ and $s_{t+1}$ are independent conditional on $(s_t, a_t)$. This means that the CGM $\mathcal{G}_{r, s'} (s,a)$ can be decomposed into the `$d$-separable' sub-graphs, $\mathcal{G}_{r} (s,a)$ and $\mathcal{G}_{s'} (s,a)$, that can be treated independently. A useful measure for identifying and quantifying the strength of a causal relationship between two variables $X$ and $Y$ in a CGM is the Causal Information Flow:
\begin{definition}[Causal Information Flow \citep{janzing2013quantifying}]
    Let $X$ and $Y$ be random variables in a causal graph, $X, Y \in \mathcal{V}$, and $pa(Y)$ be the parents of $Y$. The \textbf{Causal Information Flow} from $X$ to $Y$ is defined as: $I(X \rightarrow Y) := I(X;Y \mid pa(Y) \backslash \{X\} ) $, where $I(X;Y)$ is the mutual information --- $I(X;Y) = H[X] - H[X|Y]$ --- between $X$ and $Y$, and $I(X;Y|pa(Y) \backslash \{X\})$ is the conditional mutual information given $pa(Y)$ excluding $X$.
\end{definition}
Causal Information Flow is helpful not only to identify the presence of an edge $X \rightarrow Y$, but also to quantify the ``causal strength'' between the two nodes in a graph $\mathcal{G}$: if $X \not\rightarrow Y$ then $I(X \rightarrow Y) := I(X;Y|pa(Y) \backslash \{X\}) = 0$, otherwise $I(X \rightarrow Y) := I(X;Y|pa(Y) \backslash \{X\}) > 0$. Note that a Causal Information Flow equal to zero, $I(X \rightarrow Y) = 0$, implies the conditional independence $Y \perp X \mid pa(Y) \backslash \{X\}$. Using the notion of causal information, or equivalently of conditional independence, we can derive the following result regarding C-MDPs:   
\begin{proposition}[Identifiability of Local CGMs] \label{prop1}
Consider a C-MDP $\mathcal{M}_c = (\mathcal{M}, \mathcal{B}_{r, s'})$ with states $S = (S_1,...,S_D)$ and actions $A = (A_1,...,A_K)$. Let $\mathcal{V} = \{S_1, ..., S_D, A_1, ..., A_K\}$. Then, the following statements hold:
\begin{enumerate}[label=(\roman*), left=-5pt, labelsep=5pt, itemsep=0pt, topsep=0pt]
    \item There is no edge from vertex $V \in \mathcal{V}$ to $S'_j$, or $R$, if $\,V \perp S'_j \mid pa(S'_j) \backslash \{V\}$, or if $V \perp R \mid pa(R) \backslash \{V\}$, under any policy $\pi(s)$
    
    \item If there is no unobserved vertex $U$ such that $U$ has paths directed to both $V$ and $Y$, where $V \in \mathcal{V}$ and $Y \in \{S'_j, R\}$, then the causal edges in the local CGMs $\mathcal{G}_{s'} (s,a)$ and $\mathcal{G}_{r} (s,a)$ are all identifiable from observed trajectories.
\end{enumerate}
\end{proposition}
\vspace{-0.28cm}
\begin{proof}[Proof Sketch]
If $\,V \perp S'_j \mid pa(S'_j) \backslash \{V\}$ holds, then by the Markov property (and under faithfulness) there are no other unblocked path from $V$ to $S'_j$ in the graph $\mathcal{G}_{s'}$, except for the ones associated with other $pa(S'_j)$ (the same holds for $\mathcal{G}_{r}$ with $R$). Since the `backward' path to previous states is blocked by Markov temporal constraints, then $V \rightarrow Y \in \{S'_j, R\}$ is the only path except other identified direct $pa(S'_j)$. Note that this holds for any policy $\pi(s)$ since it holds for any intervention where $do(A=a), \, \forall a \in \mathcal{A}$. As for \emph{(ii)}, if there were an unobserved vertex $U$ such that $V \leftarrow U \rightarrow Y \in \{S'_j, R\}$, this would create a non-causal path $V \leftrightarrow Y \in \{S'_j, R\}$ that cannot be blocked via conditioning, implying $I(V;S'_j|pa(S'_j) \backslash \{V\}) > 0$ even in the case where $V \not\rightarrow Y$.
\end{proof}
\vspace{-0.28cm}
Proposition \ref{prop1} is useful in two ways. The first part tells us how we can reliably detect an edge in the C-MDP's local CGMs. The second part instead tells us that, under the assumption of no unobserved confounders $U$, we can safely identify the full local CGMs' structure via the transitions gathered by the agent and recover it by using, e.g., conditional independence testing methods. Given this result, the goal of this work is to propose a causally robust and generalizable MBRL algorithm (C-MBPO) that can reliably learn the task described by the Causal MDP. C-MBPO harnesses the (approximate) structural causal models underlying $\mathcal{G}_{r} (s,a)$ and $\mathcal{G}_{s'} (s,a)$, i.e., the collections of functions $s'_d = f_d(pa(s'_d))$, $\forall d \in \{1, ..., |S|\}$ and $r = f_r( pa(r))$, to generate `pseudo' on-policy trajectories in a DynaQ and MBPO fashion \citep{janner2019trust}. We will show how embedding causal structure into MBRL agents delivers policies that are robust to shifts affecting spurious variables.

\section{Causal Model-Based Policy Optimization}

Consider an agent that faces the following task. They pull a lever that operates a door. They observe, as part of the state space, the lever state and a light that goes on when the door is open. They receive high rewards when the door is successfully opened. The light state is merely an indicator of the door state, and does not affect rewards. A non-causal agent is likely to infer that the spurious relation $\texttt{Light state} \rightarrow \texttt{Reward}$ hold, in addition to the true causal one $\texttt{Light state} \leftarrow \texttt{Door state} \rightarrow \texttt{Reward}$, and thus learn a policy $\pi(s_t)$ assuming they have control over $\texttt{Light state}$. The true causal mechanism can be recovered by conditioning on $\texttt{Door state}$, which makes $\texttt{Light state}$ and $\texttt{Reward}$ conditionally independent: $\texttt{Reward} \perp \texttt{Light state} \mid \texttt{Door state}$. The same holds for $A$ and $\texttt{Light state}$: $A \perp \texttt{Light state} \mid \texttt{Door state}$. More practically, suppose the state space is $S_t = (X_t, Z_t) \in [-1, 1]^2$, action is $A_t \in [-1, 1]$, and rewards are $ R_t \in [-1, 1]$. The CGM associated with this problem is depicted in Figure \ref{fig:exp}a). The red dashed lines indicate that both states can be factored in as policy inputs. The SCM is defined as:
\begin{gather} \label{eq:SCM}
    X_{t+1} = f_X (X_t, A_t) + \varepsilon_x ~ , \quad \varepsilon_x \sim \mathcal{N} (0, 0.1^2) ~ , \nonumber  \\
    Z_{t+1} = f_Z(X_{t+1}) + \varepsilon_z ~ , \quad \varepsilon_z \sim \mathcal{N} (0, 0.1^2) ~ , \\
    A_t \sim \pi_{\psi}(X_t, Z_t) ~, \quad R_t = f_r(X_t) + \varepsilon_r ~ , \quad \varepsilon_r \sim \mathcal{N} (0, 0.1^2) ~ , \nonumber
\end{gather}
where the policy $\pi_{\psi}(X_t, Z_t)$ is parametrized by some $\psi \in \Psi$. The functionals $f_j (\cdot)$ are either simple linear or non-linear functions. Here, $X_t$ is the relevant variable in the system's state: $X_t$ can be modified by $A_t$ directly and is the only parent of $R_t$. Variable $X_t$ is a pure mediator for the causal effect of $A_t$ on $Z_t$, and a confounder for the effect of $Z_t$ on $R_t$. A model-free agent relies only on transitions gathered from the environment to update $\pi_{\psi} (s_t)$. A MBRL agent, such as MBPO (Model-Based Policy Optimization) \citep{janner2019trust}, instead uses transitions gathered from the environment to learn an approximate probabilistic model of the dynamics $p (r_t, s_{t+1} | s_t, a_t) = p (r_t | s_t, a_t) p (s_{t+1} | s_t, a_t)$, via an ensemble of $M$ probabilistic neural networks $\{f_m (s_t, a_t ; \theta_m) \}^M_{m=1}$, to better capture uncertainty in the transitions. Then a combination of model-generated trajectories and environment's real trajectories are used to update $\pi_{\psi} (s_t)$ in a data-augmentation fashion. This allows incorporating pseudo on-policy data, by using the model to rollout ``imaginary'' trajectories from past states under the current $\pi_{\psi} (s_t)$, making MBPO highly sample efficient. At the same time however, small errors in the model can compound quickly when rollout's length increases (and typically, reliance on environment-collected trajectories decays to a minimum as the training progresses and the model converges). In the scenario described above, these types of non-causal agents observe a spurious association between $A_t \leftrightarrow Z_t$ and $Z_t \leftrightarrow R_t$, when in reality these are conditionally independent given $X_t$. If, at test time, the environment undergoes a distribution shift so that $Z_t$ no longer correlates with $R_t$ --- e.g., in the example above, the light stops functioning, or more in general some other external factor set the value of $Z_t$ --- the learned policy $\pi_{\psi} (s_t)$ can fail catastrophically. A causal agent that can infer that $I(A_t \rightarrow Z_t) = 0$ and $I(Z_t \rightarrow R_t) = 0$, would instead recognize that $Z_t$ can be safely ignored. The proposed approach, Causal MBPO, aims at combining tools from structure learning/causal discovery within the MBPO framework, in order to augment data with, at first, and then fully rely on, transitions generated from the approximate structural causal model. On a high level, C-MBPO is broken down into the following components:
\begin{itemize}[left=3pt, labelsep=5pt, itemsep=-1pt, topsep=-3pt]
    \item[1)] \textbf{Local CGM learning.} A batch of trajectories $\mathcal{D}^n = \{s_i, a_i, r_i, s'_i\}^N_{i=0}$ is queried by following an initial policy (random). The local CGM is estimated using conditional independence testing algorithms (e.g., PC algorithm, GES, etc.), by enforcing temporal constraints ($S^j_t \rightarrow S^j_{t+1}$ and $S^j_t \not\rightarrow S^j_{t+1}$) and/or other known constraints.

    \item[2)] \textbf{Structural Causal Model learning.} Given the learned local CGMs $\mathcal{G}_{r} (s,a)$ and $\mathcal{G}_{s'} (s,a)$, conditional distributions $p_\theta (s'_j \mid \mathrm{pa}(s'_j) )$ and $p_\theta (r \mid \mathrm{pa}(r) )$ are learned. We utilize ensembles of $M$ probabilistic neural networks $\{f_m (s_t, a_t ; \theta_m) \}^M_{m=1}$ to approximate each $p(x | pa(x))$. This allows modeling inherent uncertainty in the SCM.

    \item[3)] \textbf{Counterfactual rollouts.} We use the approximate SCM, consisting of the set of conditional distributions $p(x^j | pa(x^j))$, where $x_j \in \{s^1, ..., s^d, r\}$, to generate counterfactual transitions from current and past states $s_t$, under the current $\pi_{\psi} (s)$. Under the assumption that the agent has learned an approximately correct causal model of the environment, these transitions are faithful to the local CGM and are net of spurious relations.

    \item[4)] \textbf{Policy updates.} Early in training, a mix of real transitions $\{(s_i, a_i, r_i, s'_i)\}^N_{i=1}$ and model-based counterfactual transitions $\{(\tilde{s}_i, \tilde{a}_i, \tilde{r}_i, \tilde{s}_{i+1})\}^N_{i=1}$ are employed to update $\pi_{\psi}(s)$. In later stages, once the agent has learned a reliable local CGM and the associated SCM, $\pi_{\psi}(s)$ is updated only using the counterfactual trajectories $\{(\tilde{s}_i, \tilde{a}_i, \tilde{r}_i, \tilde{s}_{i+1})\}^N_{i=1}$.
\end{itemize}
A pseudo-algorithm of C-MBPO is depicted in \ref{alg:cmpo}. C-MBPO yields policies $\pi_{\psi}(s)$ that are robust to spurious shifts. Notice that C-MBPO represents also a more interpretable alternative to standard MBRL methods, given that it learns an explicit SCM of the environment, making the agent’s decisions traceable to specific cause-and-effect relationships.

\begin{algorithm}[t] \small
\caption{Causal Model-Based Policy Optimization (C-MBPO)}
\label{alg:cmpo}
\begin{algorithmic}[1]
\STATE Initialize policy $\pi_{\psi}$, CBN object $(\mathcal{G}_{r, s'} (s,a), \Phi)$, replay buffer $\mathcal{D}_\mathrm{env}$, and model dataset $\mathcal{D}_\mathrm{model}$
\FOR{$T$ steps}
    \STATE Collect initial data from environment under $\pi_\psi$; add $\{(s_t,a_t,r_t,s_{t+1})\}^N_{i=1}$ to $\mathcal{D}_\mathrm{env}$
    \STATE Estimate local CGMs $\mathcal{G}_r(s,a)$ and $\mathcal{G}_{s'}(s,a)$ using structure learning method --- enforce known constraints
    \STATE Initialize SCM models $p_{\theta} (x^j | pa(x^j))$ based on $\mathcal{G}_r(s,a)$ and $\mathcal{G}_{s'}(s,a)$
\ENDFOR
\vspace{0.1cm}
\FOR{$E$ epochs}
    \FOR{$N$ steps}
        \STATE Train SCM models $p_{\theta} (x^j | pa(x^j))$ for $K$ steps using $\mathcal{D}_\mathrm{env}$
        \STATE Take action according to $a \sim \pi_{\psi} (s)$; add transition to $\mathcal{D}_\mathrm{env}$
        \FOR{$M$ model rollouts}
            \STATE Sample past states $\{s_i\}_{i \leq t}$ from $\mathcal{D}_\mathrm{env}$
            \STATE Use learned SCM to perform $h$-step counterfactual rollout from $\{s_i\}_{i \leq t}$ under $\pi_{\psi}$ interventions; add to $\mathcal{D}_\mathrm{model}$
        \ENDFOR
        \vspace{0.1cm}
        \FOR{$G$ policy gradient updates}
            \STATE Update policy $\pi_{\psi}$ using SCM generated counterfactual trajectories $\mathcal{D}_\mathrm{model}$ (mix with $\mathcal{D}_\mathrm{env}$ early in training)
        \ENDFOR
    \ENDFOR
\ENDFOR
\end{algorithmic}
\end{algorithm}

\subsection{Simulated Experiment}

\begin{figure}[t]
    \centering
    \includegraphics[width=0.9\linewidth]{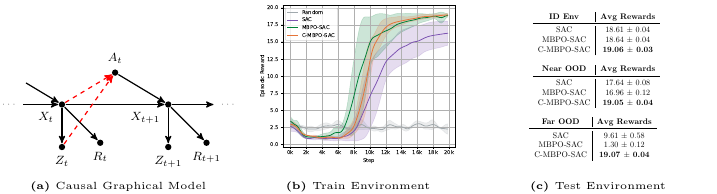}
    \caption{Figure contains, in order: a) depiction of the CGM underlying the problem; b) train-time episodic rewards for the three algorithms (SAC, MBPO-SAC, C-MBPO-SAC); c) average episodic rewards (plus 95\% intervals) of the three methods, over $B=100$ episodes, on the In-Distribution (ID) environment they were trained on, the Near Out-of-Distribution (Near OOD) and the Far OOD ones where we have intervened on $do(Z_t=z)$.}
    \label{fig:exp}
\end{figure}

Going back to the simple environment presented in earlier sections, where $S_t = (X_t, Z_t) \in [-1, 1]^2$, $A_t \in [-1, 1]$, and $ R_t \in [-1, 1]$, and with true SCM described by the set of equations in \ref{eq:SCM}, we carry out a comparison between the following different agents: i) a model-free SAC agent; ii) a MBPO-SAC agent; iii) a C-MBPO-SAC agent, as described above. The agents are trained over 100 episodes, with episode length $T=200$. The reward function is defined as $R_t = 0.1 - 0.1 X_t^2 + \varepsilon_{r,t}$. Figure \ref{fig:exp}b) reports episodic rewards $\sum^T_t r_t$ of the three methods during training. All methods converge to high reward regions reasonably quickly, with MBPO-SAC and C-MBPO-SAC reaching high rewards quicker due to their ability of leveraging model generated, pseudo on-policy trajectories to update $\pi_{\psi}$. After training all three agents on the environment, we validate them at test time by deploying the learned policies $\pi_{\psi}$ on three different versions of the environment: i) the In-Distribution (ID) version of the environment they were trained on. ii) A Near Out-of-Distribution (Near OOD), version of the environment, where we intervene on $Z_t$ by setting it to $do(Z_t = 0), \, \forall t$. In the motivating example above, this would be equivalent to the light breaking, i.e., $\texttt{Light state} = 0$. iii) Lastly, a Far OOD version of the environment, where the functional relationship in $X_t \rightarrow Z_t$ is inverted from $Z_t = X_t + \varepsilon_{z,t}$ to $Z_t = - X_t + \varepsilon_{z,t}$. Results are shown in Table \ref{fig:exp}c), where we report episodic rewards, averaged across 100 episodes, on the ID, Near OOD and Far OOD versions of the environment. In the ID table we observe that all three agents achieve similar performance to training. In the Near OOD case, performance of the non-causal agents, SAC and MBPO-SAC, starts to significantly deteriorate, while C-MBPO-SAC display statistically equivalent performance to the ID scenario. Finally, in the Far OOD scenario, SAC and MBPO-SAC display even worse performance, with MBPO-SAC's policy failing catastrophically due to its dual reliance on spurious correlations. Indeed not only does MBPO-SAC base its policy on observational data that wrongly suggest that $Z_t \rightarrow R_t$ (and $A_t \rightarrow Z_t$), as in model-free SAC, but it also explicitly encodes this spurious dependency when learning the reward model $p_{\theta}(R_t | X_t,Z_t)$, further compounding model error when rolling out `imaginary' trajectories under distributional shifts. C-MBPO-SAC instead, has correctly identified that $Z_t$ is a mere sink variable, and that the agent exercises no direct control over it, once conditioning on $X_t$.  

\section{Conclusion}

The C-MBPO method presented in this paper explicitly leverages the causal structure underlying an environment's dynamics, enabling agents to learn policies that are both generalizable and robust to distribution shifts, as demonstrated in the simulated experiment. By learning a local SCM, C-MBPO also enhances interpretability, by allowing practitioners to trace which variables drive decisions, and how. Future work on the topic will focus on refining the way the CGM is learned, by introducing a belief state over $\mathcal{G}$ that allows for general uncertainty around the CGM.

\setlength{\bibsep}{2.5pt} 
\bibliography{Refs}
\bibliographystyle{apalike}

\end{document}